\documentclass{article}

\usepackage[final,nonatbib]{nips_2016}
\usepackage[utf8]{inputenc} 
\usepackage[T1]{fontenc}    
\usepackage{hyperref}       
\usepackage{url}            
\usepackage{booktabs}       
\usepackage{amsfonts}       
\usepackage{nicefrac}       
\usepackage{microtype}      
\usepackage{amsmath,graphicx}
\usepackage{subfigure}
\usepackage{pgfplots}
\usepackage{multirow}

\title{Efficient Convolutional Neural Network\\with Binary Quantization Layer}

\author{Mahdyar Ravanbakhsh$^{1}$\quad Hossein Mousavi$^{2}$\quad Moin Nabi$^{3}$\quad Lucio Marcenaro$^{1}$
	\\\textbf{Carlo Regazzoni$^{1}$}\\ 
  $^{1}$DITEN, University of Genova\quad $^{2}$PAVIS, Istituto Italiano di Tecnologia\\
  $^{3}$DISI, University of Trento \\
  \texttt{mahdyar.ravan@ginevra.dibe.unige.it,\;hossein.mousavi@iit.it} \\
  \texttt{moin.nabi@unitn.it,\;lucio.marcenaro@unige.it,\;carlo.Regazzoni@unige.it}\\
}

\begin{document}

\maketitle

\begin{abstract}
  In this paper we introduce a novel method for segmentation that can benefit from general semantics of Convolutional Neural Network (CNN). Our segmentation proposes visually and semantically coherent image segments. We use binary encoding of CNN features to overcome the difficulty of the clustering on the high-dimensional CNN feature space. 
   These binary encoding can be embedded into the CNN as an extra layer at the end of the network. This results in real-time segmentation. To the best of our knowledge our method is the first attempt on general semantic image segmentation using CNN. All the previous papers were limited to few number of category of the images (e.g. PASCAL VOC). Experiments show that our segmentation algorithm outperform the state-of-the-art non-semantic segmentation methods by large margin.
\end{abstract}

\section{Introduction}
\label{sec:intro}
Image segmentation is a challenging task in computer vision that can identify the visual elements in an image. These elements can be used as the building blocks for any image understanding method. Traditionally, these image segments are optimized to be semantic (e.g. be an object, part of an object, or part of a scene) and visually coherent; This means that nearby pixels in each segment must have similar intensity \cite{felzenszwalb2004efficient,achanta2012slic,khoreva2016improved}. Semantic image segmentation has been proposed in several articles \cite{girshick2014rich,long2015fully,chen2014semantic,zheng2015conditional,noh2015learning}. All of these methods are limited to a narrow scope of semantics. They can only find the segments belong to \emph{few categories} of objects (e.g. 20 categories in PASCAL VOC dataset). In this paper a method is proposed that can find \emph{general} semantic segments. 

Recently there has been a remarkable progress in computer vision through Deep Neural Networks. More specifically, with Convolutional Neural Networks (CNNs) end-to-end object recognition modules have been created \cite{alexnet, simonyan2014very, szegedy2015going} outperforming all of the previous recognition systems. 
Moreover, \cite{donahue2013decaf} showed that these features are so powerful that can be used for a variety of tasks in computer vision. Given an image as input we can apply a fully-convolutional neural network to obtain a feature vector per each receptive-field in the image\cite{long2015fully}. Since these features carry semantical information about the input image, they can be used to find image segments that are semantically coherent. In this paper, we show how these segments can be extracted from such CNN features.            

CNN features are very high-dimensional (namely, 4096). Traditional segmentation approaches that are mainly based on clustering techniques \cite{shi2000normalized} are not feasible. Since we want each segments corresponds to a meaningful visual element, large number of cluster centers are essential. That makes the segmentation process even more complex. To overcome such computational complexities, binary encoding of CNN features is proposed instead. A CNN feature is converted to a short binary code: each bit pattern represents a cluster center in the original CNN feature space. For example, a 32-bit binary code can generate $2^{32}$ clusters. Each bit corresponds to a visual attribute. Nearby pixels should have similar binary patterns unless they undergo a large semantical change. This is a perfect property to be used for semantic segmentation. Iterative-Quantization (ITQ) \cite{gong2013iterative} is employed to learn these binary codes. An Interesting property of the ITQ is that it generates bits in a simple way and the transformation is linear. This is a perfect setting to be embedded in the CNN networks as a new layer. 
Once the binary map of the CNN features is available, a low-level superpixel extraction method is applied on the whole image and then the superpixels with the similar binary patterns (under Hamming distance) are merged together.


Major contributions in this work can be summarized as; \emph{1)} a semantic segmentation is proposed which can be used in a general setting, unlike the all previous methods that are limited to specific categories. \emph{2)} a compact representation of high-dimensional CNN features is introduced in the form of binary codes, to preserve semantic information, thus it can be used for semantic segmentation. Hence, we present a binary encoding layer in our network, which can be updated using back-propagation. This new layer is able to be attached to any other deep-net for encoding purposes. The primarily version of this paper has been published in~\cite{cnnseg}. 

\section{Related Work:}
\label{sec:Related}
Despite a large number of works on low-level segmentation, there are few works targeting semantic segmentation, and to the best of our knowledge, there is no work doing \emph{general} semantic segmentation using high-level CNN features.\\
\noindent\textbf{1) Low-level segmentation:} 
 refers to partitioning an input image into a set of perceptually meaningful atomic regions, considering the low-level image features, like intensity, edge, or texture. 
 In literature, apart from the core low-level feature used, a substantial debate has been mainly posed over the optimization algorithms employed to efficiently solve this partitioning problem. In this context, two classes of approaches can be identified~\cite{achanta2012slic}.
On one hand, \emph{graph-based} methods treat pixels as nodes in graph, connected each other via edges reflecting their similarity in the feature space. Then, the graph is partitioned into a set of sub-graphs corresponding to image segments by minimizing a cost function. 
Among the best performing methods, Normalized-Cuts~\cite{shi2000normalized}, Super-pixel Lattices~\cite{moore2008superpixel}, and Efficient Graph-based Segmentation (EGS)~\cite{felzenszwalb2004efficient} can be quoted. Here in this work, our method is compared with the last work, selected as one of the best performing non-semantic graph-based segmentation methods. On the other hand, a different set of methods, named \emph{gradient-ascent-based} approaches, starts with an initial clusters of pixels, then refines iteratively until convergence in visual consistency. In this line of research, Mean-shift \cite{comaniciu2002mean}, Turbo-pixel \cite{levinshtein2009turbopixels} and state-of-the-art SLIC \cite{achanta2012slic} should be mentioned. Note that, SLIC has picked as one of our baselines and compare our method with.\\
\noindent\textbf{2) Semantic segmentation:} 
Recently, visual recognition task has come to rely increasingly on segmentation, and region extraction, accordingly, emerged to play a key role in object detection \cite{girshick2016region} and activity recognition\cite{jain2014action,gkioxari2015finding,plugnplay}. 
Semantic segmentation often formulated as combining low-level segments with region-based object detectors either in a cascade \cite{arbelaez2012semantic,carreira2012object,van2011segmentation} or joint \cite{dong2014towards,mottaghi2014role,yao2012describing} manner. Convolutional Neural Networks have recently resurfaced as a powerful tool for learning to segment semantically \cite{girshick2014rich,long2015fully,zheng2015conditional,chen2014semantic}. Nevertheless, learning such supervised deep structures for higher number of categories (and samples) is so supervision-demanding and computationally-expensive. Very recently, ADE20K~\cite{zhou2016semantic} has been introduced in which a wider variety of scenes and objects are annotated. Even in this case, extending the current supervised DNNs to work in a zero-shot fashion (namely, the categories other than the ones exist in the dataset) is not trivial.\\
In this work, however, a completely different perspective to semantic segmentation is picked out. We specifically propose a method to narrow down the semantic gap (between pixels and concepts) in images, namely, trying to inject semantic inherited from generic CNN representations, so leading to more general semantic segmentation while maintaining the method complexity to a manageable level.

\begin{figure}[htb]
	\centering
	
	\includegraphics[width=\linewidth]{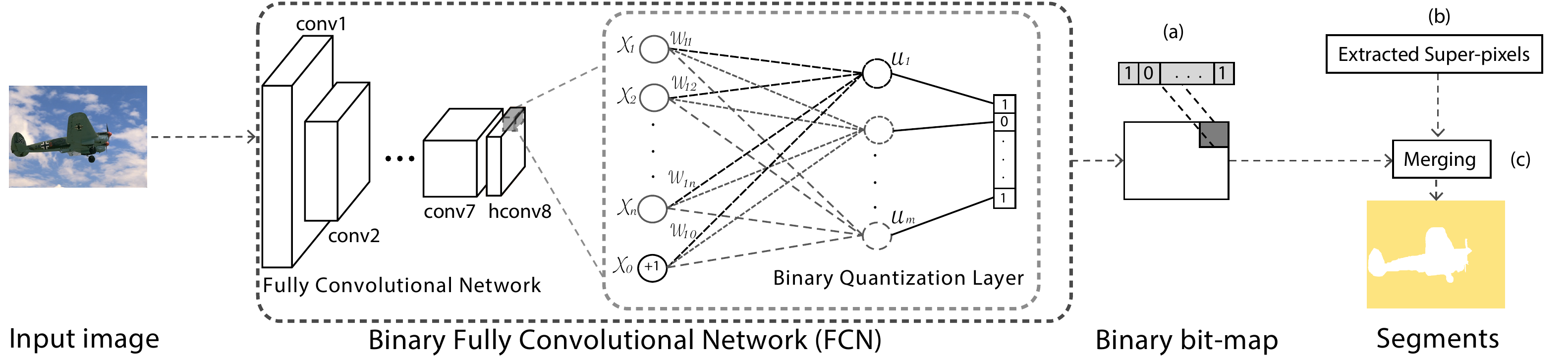}
	\caption{\small Method overview: given an image forward-pass through the net (a), a low-level superpixel is extracted (b), the binary code of each superpixel is assigned to merge the superpixels with similar binary patterns (c).}
	\label{fig:overview}
\end{figure}


\section{CNN-aware binary map of image segments}
\label{sec:Method}
In this section, the three major parts of our proposed method are described in details: {\it 1) Spatial-aware fully convolutional network}, {\it 2) Binary map encoding layer} and {\it 3) Semantic segmentation using binary maps}. Figure~\ref{fig:overview} illustrates a general work-flow of the method.\\
\noindent\textbf{Spatial-aware Fully Convolutional Network:}
Early convolutional layers in CNNs represent more local information of the image, while deeper ones contain more global information. The fully-connected layers capture higher-level information and are usually employed for recognition purposes. It has been shown that the deep nets which trained on ImageNet~\cite{deng2009imagenet}, are rather semantic; they can address wide range of recognition problems~\cite{razavian2014cnn,donahue2013decaf}. Fully convolutional Nets also can preserve relative spatial coordinates between input image and output feature map. These properties motivated us to use such structures for general semantic segmentation.\\
\noindent\textbf{Binary Encoding Layer:}
Clustering the high-dimensional feature maps comes with high computational cost. It leads to converge to a limited number of clusters. One possible solution to avoid this problem is partitioning high-dimensional features into a set of buckets (instead of clusters) using hashing techniques. 
Obviously, dealing with binary codes comes with lower computational cost and higher efficiency with respect to other clustering methods. However, the most advantage of hashing comparing to clustering, is the capability of embedding it simply as a layer inside the network.\\
Encoding feature maps to binary codes is done by ITQ\cite{itq}, which is a hashing method for unsupervised-learning binary codes. Training the ITQ is the only training cost in the proposed method, which can be done only once on a subset of data.The ITQ projects each high-dimensional feature vector into a binary space. We use the hashing weights, which are learned by ITQ, to build up a binary encoding module as a last layer (denoted by $hconv8$) in the network architecture. 
We implement this layer as a set of convolutional filters (shown in different colors in Fig \ref{fig:overview}), followed by a sigmoid activation function. The number of these filters is equal to the size of the binary code and the weights are computed through ITQ. Finally, the binarization step has been done externally by thresholding the output of the sigmoid function.\\
Specifically, for $X=\{x_{1},x_{2},...,x_{n}\}$ a feature vector of $fc7$, the output of $hconv8$ is defined by $hconv8(X) = XW_{i}$, where $W_{i}=\{w_{i1},w_{i2},...,w_{in}\}$ are the weights for the $i^{th}$ neuron. The non-linearity is provided with a sigmoid function $\upsilon= \sigma(hconv8(X))$, and the threshold function is:
\begin{equation}
\label{eq:activation}
g(\upsilon) = \Big\{
\begin{tabular}{cc}
0, & $\upsilon \le 0.5$\\
1, & $\upsilon > 0.5$
\end{tabular}
\end{equation}
Such a binary quantization layer can be plugged into the net as a pre-trained module, with the possibility of fine-tuning with back-propagation in an end-to-end fashion.\\
\noindent\textbf{Semantic Segmentation Using Binary Maps:}
The generated binary maps have two important aspects: {\it 1)} it preserves the spatial relation between input image and output features. In other word, each region on the binary map corresponds to a patch on the input image; {\it 2)} binary maps are generated using the convolutional feature maps of the deep-net, hence they capture the semantics of the scene. Binary patterns with different values represent areas with different semantics. It is specifically interesting because, any changes in the binary code patterns on binary maps can be interpreted as a semantics change on the corresponding areas on the image. We take advantage of these two properties, first the segmentation initialized by low-level superpixels, then merged superpixels with the similar binary codes in the binary map. This simple yet effective criteria on semantic features has shown to be much more powerful compared to the previous state-of-the-art methods which adopted much more sophisticated partitioning algorithms but relayed only on low-level visual information.
\begin{figure}
	\centering
	\small{Original Image \; \; \; \;  Ground truth \; \;  \; \; \; \; \; \;  EGS \; \; \; \; \; \; \; \; \; K-means \; \; \; \; \; Our Method\;}
	
	\includegraphics[width=2.4cm,height=1.2cm]{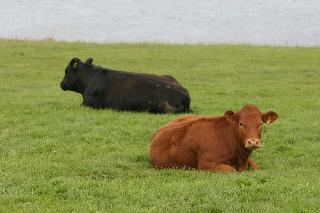}
	\includegraphics[width=2.4cm,height=1.2cm]{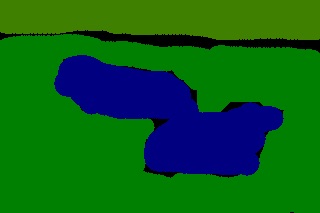}
	\includegraphics[width=2.4cm,height=1.2cm]{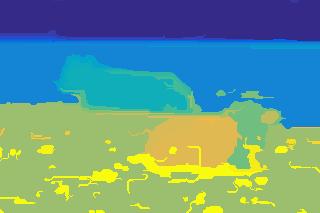}
	\includegraphics[width=2.4cm,height=1.2cm]{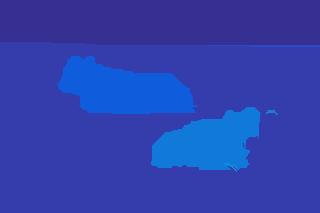}
	\includegraphics[width=2.4cm,height=1.2cm]{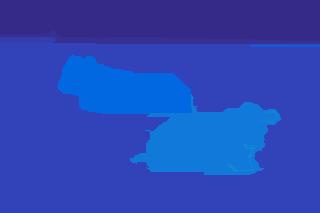}
	
	\includegraphics[width=2.4cm,height=1.2cm]{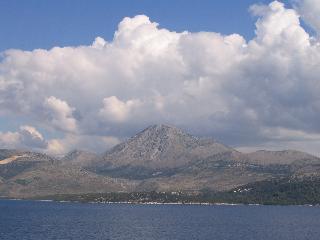}
	\includegraphics[width=2.4cm,height=1.2cm]{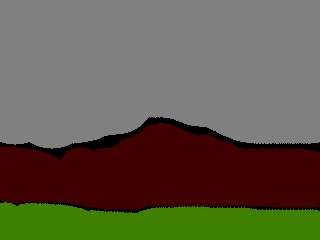}
	\includegraphics[width=2.4cm,height=1.2cm]{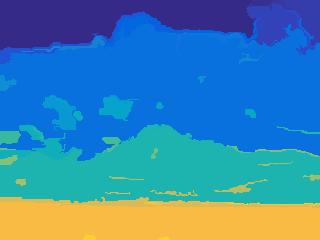}
	\includegraphics[width=2.4cm,height=1.2cm]{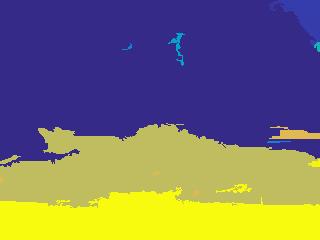}
	\includegraphics[width=2.4cm,height=1.2cm]{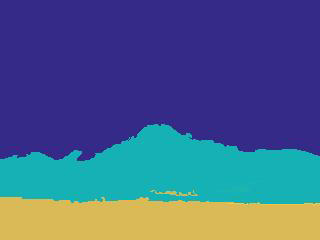}

	\includegraphics[width=2.4cm,height=1.2cm]{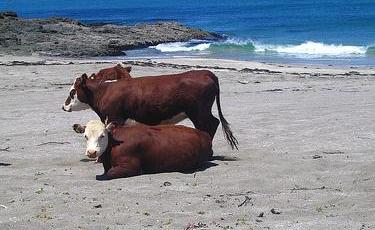}
	\includegraphics[width=2.4cm,height=1.2cm]{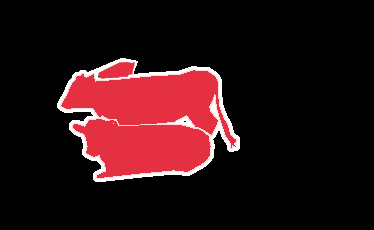}
	\includegraphics[width=2.4cm,height=1.2cm]{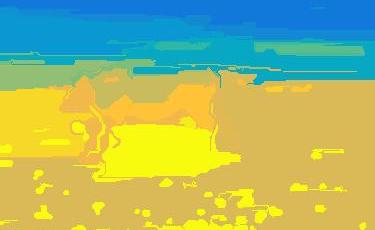}
	\includegraphics[width=2.4cm,height=1.2cm]{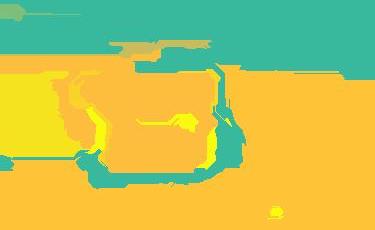}
	\includegraphics[width=2.4cm,height=1.2cm]{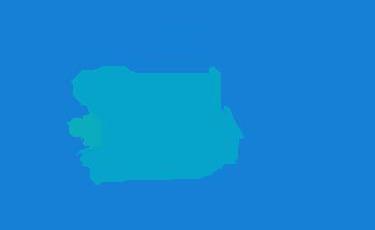}
	
	\includegraphics[width=2.4cm,height=1.2cm]{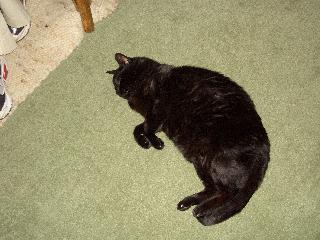}
	\includegraphics[width=2.4cm,height=1.2cm]{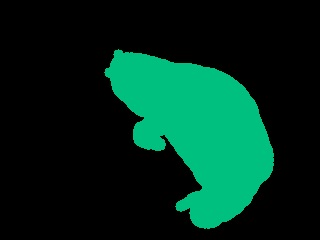}
	\includegraphics[width=2.4cm,height=1.2cm]{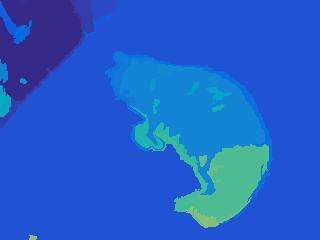}
	\includegraphics[width=2.4cm,height=1.2cm]{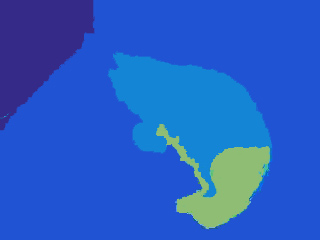}
	\includegraphics[width=2.4cm,height=1.2cm]{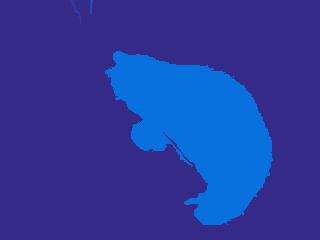}
	
	\caption{\small{Our method compared with Efficient Graph-based Segmentation (EGS) \cite{felzenszwalb2004efficient} and k-means clustering}}
	\label{fig:db}    
\end{figure}

\section{Results}
\label{sec:Experiment}
\noindent\textbf{Experimental Setup:}
Several comparative experiments has been done to demonstrate the advantage of using our binary map for semantic segmentation compared to two best-performing low-level segmentation methods: \emph{Efficient Graph-Based Segmentation (EGS)}\cite{felzenszwalb2004efficient} and \emph{gradient-ascent-based SLIC}\cite{achanta2012slic}.
Also, a clustering-based method is selected as a baseline. The K-means clustering method is applied to merge the super-pixels. 
Since there is no pre-processing step nor parameter selection in our method, for the baselines, we used the publicly available codes with the default parameters ($\sigma = 1.0, k = 100$). The main objective of this experiment is to show the strength of the semantic segmentation compared to low-level segmentation without any parameter tuning. 
Two datasets were considered; the Berkeley Segmentation Dataset (BSDS500), and Microsoft Research Cambridge database (MSRC). Both evaluations have been performed with the original parameters' setups for these datasets.\\
\noindent\textbf{Evaluation:} we adopted the Segmentation Intersection over Union (IoU) as one of the most commonly used evaluation measure for segmentation task. IoU is defined as:
\begin{equation}
\label{eq:iou}
IoU(P_{m},P_{gt}) = \frac{|P_{m} \bigcap P_{gt}|}{|P_{m} \bigcup P_{gt}|}
\end{equation}
 Where $P_{gt}$ is ground truth segment annotation, and $P_{m}$ is predicted segment. As the predicted segments, we select the segments with the maximum IoU with each segment in $P_{gt}$. The final value of \emph{Segmentation-IoU} is computed as the average over all the segments of all the images of the dataset.\\
\noindent\textbf{Segmentation Network Details:}
The designed deep-net consist of two major parts; \emph{1) Fully convolutional network:} 
at first we used a pre-trained AlexNet model on ImageNet~\cite{alexnet}. Original AlexNet, contains 5 convolutional layers and two fully connected layers. In order to obtain spatial-aware feature maps, we convert the last two fully connected layers into convolutional layers. By transforming fully connected layers into convolutional layers we could enable the net to output a multi dimensional feature map disregard to input image size and produce an efficient model for spatial-aware patch pooling.
\emph{2) Binary Bit-map layer:}
Bit-map layer is designed for convolutional feature map quantization. In order to build the layer, we first extract convolutional maps from $conv7$ over PASCAL 2007 images to train an unsupervised ITQ hash to model 4096 dimensional feature maps into 8-bits binary codes. Hashing weights obtain from ITQ applied into a Depth Normalization Layer. We embedded the Depth Normalization Layer with pre-trained weights to the network to build {\it Binary Bit-map layer}. Output of the network is a set of 8-bit binary maps.\\ 
\noindent\textbf{Segmentation Strategy:}
For segmentation, we first extract the binary maps for each input image. For each superpixel a binary code is assigned by the corresponding region on binary map. Then we merged the superpixels with the similar binary codes on the bit maps (i.e., zero distance in Hamming space). The final segmentation is obtained as the result of such merging of superpixels.\\
\noindent\textbf{Results:}
our proposed semantic segmentation significantly outperform previous low-level segmentation methods as well as clustering the super-pixels by k-means method. Figure \ref{fig:db} shows the qualitative results of our method and EGS method. In compare with EGS our approach achieved to better segmentation. A comparison results on two datasets demonstrated in Table \ref{tab:compare}. The first row shows a comparison of average segmentation-IoU for the algorithms in the MSRC dataset, and the second row compares the algorithms on BSDS500. We outperform both baseline methods by large margins in term of $segmentation-IoU$ over different superpixel sizes (Figure \ref{tab:segacc}). Such evaluation shows the robustness of the proposed method to the number of super-pixels.We observe that segmentation on images containing ``things'' (objects) are significantly better as compared to images containing ``stuffs'' (scenes). It also supports our hypothesis that binary patterns preserved semantic information and the understanding objects in the scene.


\section{Conclusion}
\label{sec:conclusion}
In this work a novel approach to general semantic-aware image segmentation has been presented which does no require category-specific training a deep-net. We employed AlexNet as pre-trained model and convert fully connected layers into convolutional layers. An efficient ITQ hashing layer is attached as the final layer to the net to quantity high dimensional feature maps in form of binary code representation. Such model provides both spatial consistency as well as low dimensional semantic embedding. Our experimental results shown using these binary maps can improve the performance of the segmentation comparing to several low-level segmentation methods. 
\begin{figure}
	\centering
	\scalebox{.50}{
		\subfigure[Segmentation-IoU on Berkeley] {
			\begin{tikzpicture}[thick,scale=0.95]
			\begin{axis}[
			xlabel={Number of Super-pixels},
			ylabel={Segmentation accuracy \%},
			xmin=90, xmax=510,
			ymin=30, ymax=60,
			xtick={100,200,300,400,500},
			ytick={30,35,40,45,50,55,60},
			legend style={at={(1,1)},anchor=north east},
			ymajorgrids=true,
			grid style=dashed]
			\addplot+[
			color=blue,
			mark=square,
			]
			coordinates {
				(100,38.7)(200,48.35)(300,42.5)(400,44.78)(500,44.35)
			};
			\addplot+[
			color=red,
			mark=square,
			]
			coordinates {
				(100,34.1)(200,45.19)(300,40.4)(400,44.78)(500,42.35)
			};
			\addplot+[
			color=green,
			mark=square,
			]
			coordinates {
				(100,37.8)(200,41.7)(300,41.9)(400,43.7)(500,39.12)
			};
			\legend{Our Method, EGS Method, Slice Method}
			\end{axis}
			\end{tikzpicture}
		}
		\subfigure[Segmentation-IoU on MSRC] {
			\begin{tikzpicture}[thick,scale=0.95]
			\begin{axis}[
			xlabel={Number of Super-pixels},
			ylabel={Segmentation accuracy \%},
			xmin=90, xmax=510,
			ymin=30, ymax=65,
			xtick={100,200,300,400,500},
			ytick={30,35,40,45,50,55,60},
			legend style={at={(1,1)},anchor=north east},
			ymajorgrids=true,
			grid style=dashed]
			\addplot+[
			color=blue,
			mark=square,
			]
			coordinates {
				(100,44.5)(200,52.83)(300,55.03)(400,53.17)(500,52.1)
			};
			\addplot+[
			color=red,
			mark=square,
			]
			coordinates {
				(100,39.23)(200,47.12)(300,48.3)(400,47.3)(500,46)
			};
			\addplot+[
			color=green,
			mark=square,
			]
			coordinates {
				(100,37.77)(200,42.3)(300,45.23)(400,45.1)(500,44.9)
			};
			\legend{Our Method, EGS Method, Slice Method}
			\end{axis}
			\end{tikzpicture}
		}
	}
	\caption{\small{Segmentation-IoU over superpixel variation}}
	\label{tab:segacc}
\end{figure}
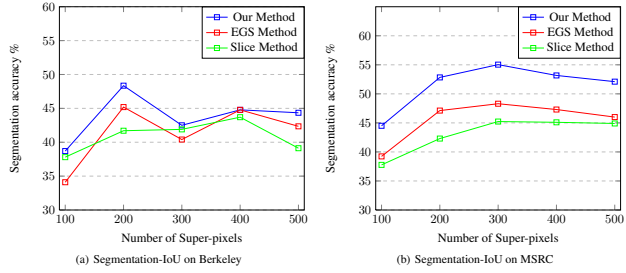
\begin{table}
	\caption{\small{Quantitative results for segmentation IoU on MSRC and Berkeley datasets.}}
	\label{tab:compare}
	\centering
		\begin{tabular}{cccccc}
			\toprule
			Dataset  & EGS \cite{felzenszwalb2004efficient} & SLIC  \cite{achanta2012slic} & k-means & Our method \\
			\midrule
			MSRC (IoU) & 50.30\% & 48.70\% & 50.80\% & \textbf{55.03 \% }\\
			\midrule
			Berkeley (IoU) & 45.19\% & 43.70\% & 44.02\% & \textbf{48.35 \% }\\
%
			\bottomrule		
		\end{tabular}
\end{table}

\small
\bibliographystyle{IEEEbib}
\bibliography{strings,refs}

\end{document}